\documentclass{bmvc2k}
\usepackage{amssymb}
\usepackage{adjustbox}
\usepackage{multirow}
\usepackage{blindtext}
\usepackage{tipa}
\usepackage{setspace}
\usepackage{todonotes}

\title{Comparing phonemes and visemes with DNN-based lipreading}

\addauthor{Kwanchiva Thangthai}{k.thangthai@uea.ac.uk}{1}
\addauthor{Helen L Bear}{helen@uel.ac.uk}{2}
\addauthor{Richard Harvey}{r.w.harvey@uea.ac.uk}{1}

\addinstitution{
 School of Computing Sciences\\
 University of East Anglia\\
 Norwich, UK
}
\addinstitution{
 Computer Science \& Informatics\\
University of East London\\
London, UK. }

\runninghead{Thangthai, Bear, Harvey}{Phonemes and visemes with DNNS}

\begin{document}

\maketitle

\begin{abstract}

There is debate if phoneme or viseme units are the most effective for a lipreading system. Some studies use phoneme units even though phonemes describe unique short sounds; other studies tried to improve lipreading accuracy by focusing on visemes with varying results. We compare the performance of a lipreading system by modeling visual speech using either 13 viseme or 38 phoneme units. We report the accuracy of our system at both word and unit levels. The evaluation task is large vocabulary continuous speech using the TCD-TIMIT corpus. We complete our visual speech modeling via hybrid DNN-HMMs and our visual speech decoder is a Weighted Finite-State Transducer (WFST). We use DCT and Eigenlips as a representation of mouth ROI image. The phoneme lipreading system word accuracy outperforms the viseme based system word accuracy. However, the phoneme system achieved lower accuracy at the unit level which shows the importance of the dictionary for decoding classification outputs into words.

\end{abstract}


\section{Introduction}
\label{sec:intro}

As lipreading transitions from GMM/HMM-based technology to systems based on Deep Neural Networks (DNNs) there is merit in re-examining the old assumption that phoneme-based recognition outperforms recognition with viseme-based systems. Also, given the greater modeling power of DNNs, there is value in considering a range of rather primitive features such as Discrete Cosine Transform (DCT) \cite{ahmed1974discrete} and Eigenlips \cite{bregler1994eigenlips} which had previously been disparaged due to their poor performance.

Visual speech units divide into two broad categories; phonemes and visemes. A phoneme is the smallest unit of speech that distinguishes one word sound from another \cite{international1999handbook}. Therefore it has a strong relationship with an acoustic speech signal. In contrast, a viseme is the basic visual unit of speech that represents a gesture of the mouth, face and visible parts of the teeth and tongue, the visible articulators.  Generally speaking, mouth gestures have less variation than sounds and several phonemes may share the same gesture so a class of visemes may contain many different phonemes. There are many choices of visemes \cite{bear2014phoneme} and Table~\ref{TB:P2V} shows one of those mappings~\cite{neti2000audio}.

\begin{table}[!htb]
\centering
\caption{Neti \cite{neti2000audio} Phoneme-to-Viseme mapping.}
\label{TB:P2V}\begin{adjustbox}{width=1\columnwidth}
\begin{tabular}{|c|rr|c|rc|c|rr|}
\hline
\multicolumn{3}{|c|}{Consonants}                                                      & \multicolumn{3}{c|}{Vowels}                                                                         & \multicolumn{3}{c|}{Silence}                                                   \\ \hline 
\multicolumn{1}{|l|}{Viseme} & TIMIT phonemes      & \multicolumn{1}{r|}{Description} & \multicolumn{1}{l|}{Viseme} & TIMIT phoneme                      & \multicolumn{1}{r|}{Description} & \multicolumn{1}{l|}{Viseme} & TIMIT phoneme & \multicolumn{1}{c|}{Description} \\ \hline \hline
/A  & /l/ /el/ /r/ /y/    & Alveolar-semivowels & /V1 & /ao/ /ah/ /aa/ /er/ /oy/ /aw/ /hh/ & Lip-rounding based vowels & /S & /sil/ /sp/ & Silence \\ 
/B & /s/ /z/ & Alveolar-fricatives  & /V2 & /uw/ /uh/ /ow/ & " & & & \\ 
/C & /t/ /d/ /n/ /en/ & Alveolar & /V3 & /ae/ /eh/ /ey/ /ay/ & " & & & \\ 
/D & /sh/ /zh/ /ch/ /jh/ & Palato-alveolar & /V4 & /ih/ /iy/ /ax/ & " & & & \\ 
/E  & /p/ /b/ /m/ & Bilabial & & & & & & \\ 
/F  & /th/ /dh/  & Dental & & & & & & \\ 
/G  & /f/ /v/ & Labio-dental & & & & & & \\ 
/H  & /ng/ /g/ /k/ /w/ & Velar & & & & & & \\ \hline
\end{tabular}
\end{adjustbox}
\end{table}

\section{Developing DNN-HMM based lipreading system}

\begin{figure}[h]
			\centering
			\includegraphics[scale=0.3]{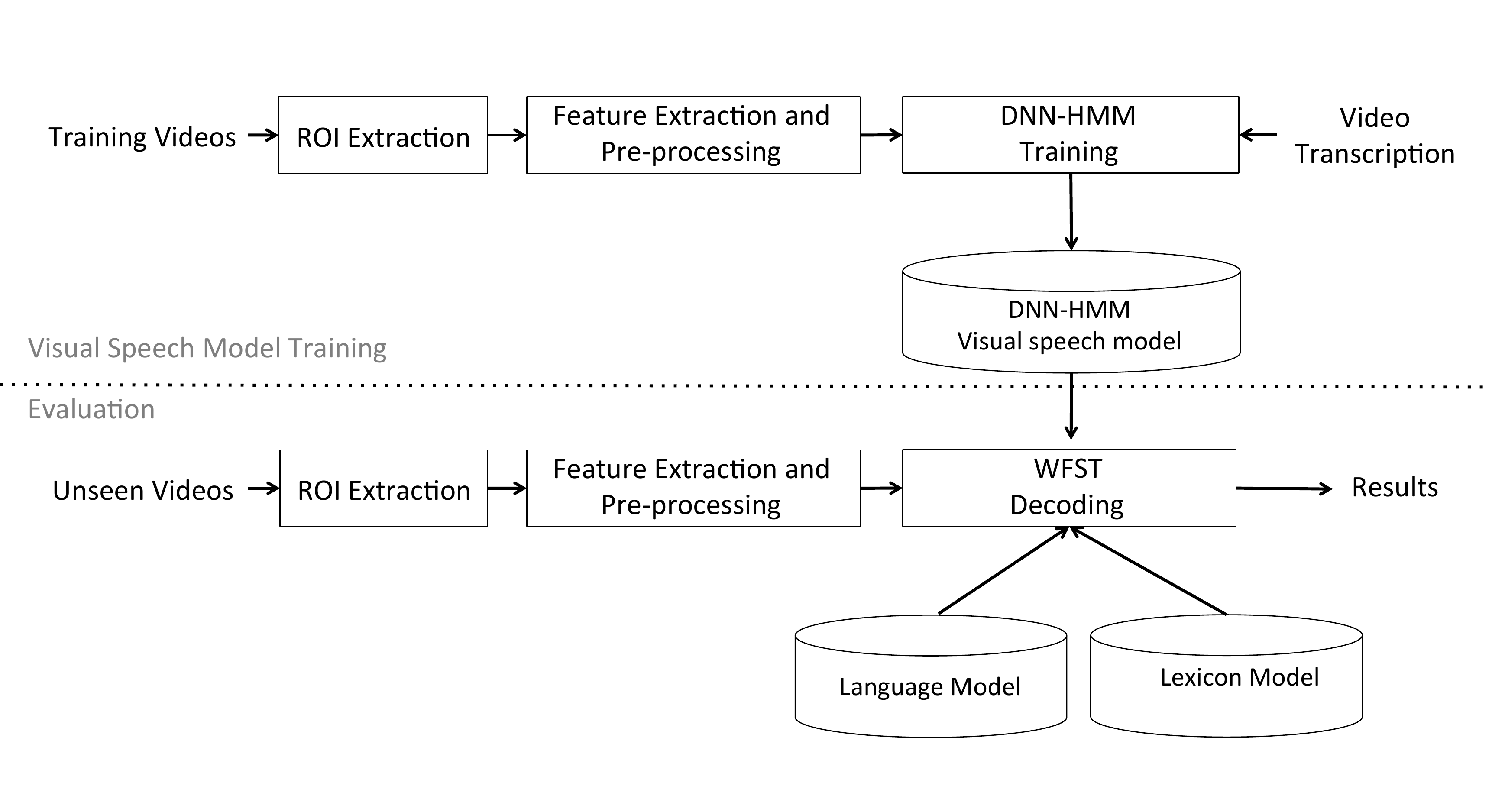} 
			\caption{Lipreading system construction techniques.}
			\label{Fig:Lip-method}
		\end{figure}

Conventional techniques to model visual speech are based around Hidden Markov Models (HMMs) \cite{rabiner1986introduction}. The aim of the model is to find the most likely word or unit sequence corresponding to the visual observation. HMMs comprise two probability distributions: the transition probability and the probability density function (PDF) associated with the continuous outputs. 
The transition probabilities represent a first-order Markov process.
The PDF of speech feature vectors is modeled by a Gaussian Mixture Model (GMM) that is parameterised by the mean and the variance of each component.

There are some weaknesses of GMM, that have been found in acoustic modeling ~\cite{Hinton2012}. First, it is statistically inefficient for modeling data that lie on or near a non-linear manifold in the data space. Second, in order to reduce the computational cost by using a diagonal rather than a full covariance matrix, uncorrelated features are needed. These deficiencies motivate the consideration of alternative learning techniques.

The deep network structure can be considered as a feature extractor by using the number of neurons in multiple hidden layers to learn the essential patterns from the input features \cite{mao1995artificial}. In addition, the backpropagation algorithm \cite{hecht1988theory} with its appropriate learning criterion is essentially optimizing the model to fit to the training data discriminatively.

However, to decode a speech signal, temporal features and models that can capture the sequential information in speech such as an observable Markov sequence in the HMM is still necessary. Thus arises the DNN-HMM hybrid structure in which the DNN is used instead of the GMM in the HMM. The method essentially combines the advantages from these two algorithms.

\subsection{Feature extraction}
\begin{figure}[!htb]
\begin{tabular}{ccc}
		\includegraphics[width=0.3 \textwidth,height=0.15\textwidth]{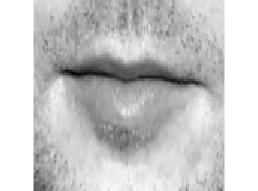} &
        \includegraphics[width=0.3 \textwidth,height=0.15\textwidth]{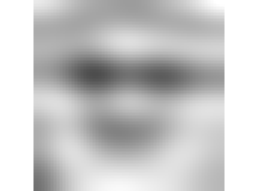} &
        \includegraphics[width=0.3 \textwidth,height=0.15\textwidth]{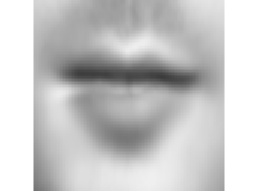} \\
\end{tabular}
			\caption{ Comparing the original ROIs image (left) and its reconstruction via $44$-coefficient DCT (middle) and $30$-coefficient Eigenlip (right).}
            \label{rois}
		\end{figure}

The literature provides a variety of feature extraction methods, often combined with tracking (which is essential if the head of the talker is moving).  Here we focus on features that have been previously described as ``bottom-up'' \cite{Matthews02} meaning that they are derived directly from the pixel data and require only a Region-Of-Interest, or ROI.  Figure~\ref{rois} illustrates a typical ROI taken from the TCD-TIMIT dataset described later in Section~\ref{sec:data} plus two associated feature representations which we now describe.

\subsubsection{Discrete Cosine Transform (DCT)}
The DCT is a popular transform in image coding and compression. DCT aims to represent the frequency domain of signal periodically and symmetrically using the cosine function. In particular, the DCT is a part of the Fourier Transform family but contains only the real part (Cosine).  Because of its popularity most modern processors execute it very quickly (roughly $\mathcal{O}(N)$ for modern algorithms) so this also explains its ubiquity. For strongly correlated Markov processes the DCT approaches the Karhunen-Loeve transform in its compaction efficiency. Possibly this explains its popularity as a benchmark feature~\cite{neti}.  Here we use DCT II with zigzag property \cite{2ddct2}, which means that the first elements of the feature vector contain the low-frequency information. The resulting feature vector has $44$ dimensions.

\subsubsection{Eigenlip}
The Eigenlip feature is another appearance-based approach\cite{KIRBY199363}. The Eigenlips feature has been generated via Principal Component Analysis (PCA) \cite{wold1987principal}.  Here, we use PCA to extract the Eigenlip feature from grey-scale lip images, where we retain only $30$-dimensions of PCA. To construct the PCA, $25$-ROIs images of each training utterance were randomly selected to be the set of training images. Almost about $100k$ images in total were used to compute the Eigenvector and Eigenvalue and use it for extracting training and testing utterances. Only $30$ dimensions of principal components with high variation were retained. 

\subsection{Feature transformation}
\begin{figure}[h]
			\centering
			\includegraphics[scale=0.14]{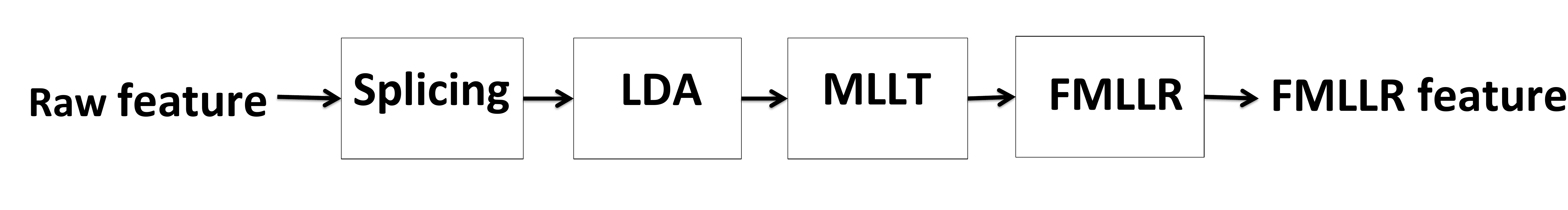} 
			\caption{FMLLR feature pre-processing pipeline.}
			\label{Fig:FMLLR}
		\end{figure}

The raw features are $30$-dimensional Eigenlip and $44$-dimensional DCT features. These raw features are then normalized by subtracting the mean of each speaker and the $15$ consecutive frames are spliced onto the feature to add dynamic information.
Second, Linear Discriminant Analysis (LDA) \cite{fisher1936use} and Maximum Likelihood Linear Transform (MLLT) \cite{Gales1998} are applied to reduce and map the features to a new space to minimize the within-class distance and maximize the between-class distance, where the class is the HMM-state, whilst simultaneously maximizing the observation likelihood in the original feature space. 

Finally, feature-space Maximum Likelihood Linear Regression (fMLLR) \cite{fMLLR2006},\cite{Gales1998}, also known as the feature-space speaker adaptation technique, is employed to normalize the variation within a speaker. These new $40$-dimensional fMLLR features are used as inputs (labeled as feature pre-processing pipeline in Figure~\ref{Fig:FMLLR}) to the subsequent machine learning.  The use of LDA is quite commonplace in lipreading and is derived in the HiLDA framework~\cite{Potamianos2001}.  MLLT and fMLLR are commonplace in acoustic speech recognition but have only recently been applied to visual speech recognition~\cite{LipreadingSAT2016} albeit on smallish datasets.

\subsection{Visual speech model training}
Our DNN-HMM visual speech model training involves all five successive stages. Here, we detail the development of the visual speech model that we employ in this work including all steps and parameters. 
		
\subsubsection{Context-Independent Gaussian Mixture Model (CI-GMM)} 
The first step is to initialise a model by creating a simple CI-GMM model. This step creates a time alignment for the entire training corpus by simply constructing a mono phoneme/viseme model that contains $3$-state GMM-HMMs for each speech unit.  

The CI-GMMs are trained on the raw features, which are DCT and PCA, along with its first and second derivative coefficients ($\Delta +\Delta\Delta$). We use $3$-state GMM-HMMs on each visual speech unit. Instead of setting the fixed number for increasing Gaussian mixture, we have set the maximum number of Gaussian to be $1000$ so that each state will keep increasing independently until their variances reach the maximum. When the training process starts, the time alignment of training data is equally segmented and updated in every iteration for the first ten iterations, then updated every two iterations until a maximum of $40$ iterations.

\subsubsection{Context-Dependent Gaussian Mixture Model (CD-GMM)}  
The context-dependent viseme models (CD-GMMs) is specified on the same feature as in CI-GMM system. Here we use tied-state of $3$-context visual speech model, where the tied-states are obtained from the data-driven approach tree-clustering~\cite{Povey11}. We have specified the maximum number of leaf nodes to be $2000$, which limits the number of states. The maximum number of Gaussians is set to $10K$. The training iterations continue until there is convergence which in practice is fewer than $35$ iterations. We realign every $10$ iterations. 

\subsubsection{CD-GMM with LDA-MLLT feature transformation} 
This training step also uses CD-GMMs, but trained on the LDA-MLLT features. The $40$-dimensional LDA-MLLT features are formed by splicing $15$ frames of the current frame (seven on the left and seven on the right) then reducing, via LDA, to $40$ dimensions per frame. This compact set of LDA-MLLT feature parameterizes to the $40$-dimension that best associates with the visual speech unit and also comprises the dynamic of visual speech over $150$ms. Again, the different set of tied-state CD-GMM has been constructed considered to the current feature. The maximum number of leaf nodes is set to $2,500$, and the total number of Gaussians is $15K$. This step utilises the equivalent number of training iterations and the realignment as those used in the previous step.

\subsubsection{CD-GMM with Speaker Adaptive Training (SAT)} 
In a Speaker Adaptive Training (SAT) system, the CD-GMM are built on an fMLLR transformation on top of LDA-MLLT features by estimating a transform for each speaker. The same training process in the preceding step is then applied on the $40$-dimensions of fMLLR feature, where the number of leaf nodes and Gaussian are identical. 

\subsubsection{Context-Dependent Deep Neural Networks (CD-DNN)}

We construct the CD-DNNs model on the hybrid DNN-HMMs architecture. The CD-DNNs are trained and optimized by minimizing frame-based cross-entropy between the prediction and the PDF target. The PDF refers to the tied-state context-dependent label, which is generated from the SAT system, that aligned every frame. The feature we adopted for all DNN training is based on LDA+MLLT+fMLLR features with mean and variance normalization. 

The CD-DNNs model is trained on six hidden layers with $2048$ neurons per layer, where we use the sigmoid non-linearity function in each neuron. The input layer is the fMLLR feature with temporally spliced $11$ consecutive frames. The model is initialized by a stacking of Recurrent Boltzman Machines (RBM) with three iterations on a single-GPU machine. The learning rate for RBM training is $0.4$ and applying L2 penalty (weight decay) at $0.0002$. The learning rate for fine-tuning has been set to $0.008$ with dropout of $0.1$. We use the minibatch-Stochastic Gradient Descent (SGD) for fine-tuning with minibatch size of $256$. We produce a development set for tuning the network by randomly selecting $10\%$ of training data. Every DNN training iteration is required to have a cross-validation loss is lower than the previous training iteration.  If a iteration is rejected then one retries with a new stochastic gradient descent parameter.  The terminating condition is that the new loss is little different from the old loss (specifically we use a difference smaller than $0.001$ of the loss as a suitable terminating condition).

\subsection{Decode lipreading with WFST Decoder}
Weighted Finite-state Transducer decoders have been increasingly used to decode speech signal in Large Vocabulary Continuous Speech Recognition (LVCSR) tasks and have also become a state-of-the-art decoder \cite{howell2016visual}. To decode a visual speech signal, we need a visual speech model, a language model, and a lexicon or as so called, a pronunciation dictionary. 

Our lipreading decoder comprises the visual speech DNN-HMM model,the TCD-TIMIT pronunciation dictionary and the
word bi-gram language model. We generate the decoding graph as a finite-state transducer (FST) via the Kaldi toolkit ~\cite{Povey11} .Beam width pruning is applied every $25$ frames where we use $13.0$ for the Viterbi pruning beam \cite{viterbi1967error} and $8.0$ for the lattice beam and the visual speech model scale is $0.1$. The lattice that contains the entire surviving path is re-scored by applying the bigram language model with the scaling factor over the range $5-15$. Only the lowest word error rates after LM re-scoring are used.

\section{Analysis of the pronunciation dictionary}
Reducing the set of speech units, such as reducing a set of phonemes to a set of visemes, reduces the discriminative power of the classification model whilst increasing the complexity of pronunciation dictionary by increasing the volume of homophonic words. This suggests that word accuracy of a viseme based system will be lower than a phoneme based system.  The counter argument is that visemes might be simpler to classify (because there are fewer of them and they are meant to be better matched to the visual signal) so there is clearly a trade-off between homopheny and unit accuracy~\cite{cox2008challenge}.

\begin{table}[h]
\centering
\caption{Example of phoneme and viseme dictionary with its corresponding IPA symbols.}
\label{tb-homophone}
\begin{adjustbox}{width=0.7\columnwidth}
\begin{tabular}{|l|c|c|c|}
\hline 
Word Entry & IPA Symbol & Phoneme Dictionary & Viseme Dictionary \\ \hline \hline
TALK	&   t \textopeno  k         	& t ao k             & C V1 H            \\ 
TONGUE 	&   t \textturnv \textipa{N} 	& t ah ng            & C V1 H            \\ 
DOG 	&   d \textopeno g       		& d ao g           & C V1 H            \\ 
DUG 	&   d \textturnv g        		& d ah g           & C V1 H            \\ 
CARE  	&   k e r				        & k eh r           & H V3 A            \\ 
WELL  	&   w e l  			       		& w eh l          	& H V3 A            \\
WHERE 	&   w e r         				& w eh r           & H V3 A            \\ 
WEAR  	&   w e r         				& w eh r           & H V3 A            \\
WHILE 	&   w ai l         				& w ay l           & H V3 A            \\ \hline

\end{tabular}
\end{adjustbox}
\end{table}

Table \ref{tb-homophone} shows examples of the homophoneme and homoviseme words that occur in the TCD-TIMIT dictionary. Figure 4 describes the homophone problem in two ways.  On the left words are binned according to how many homophones they have. Thus the column labelled ``1 occur'' is the count of all unique words, the column labelled ``2 occur'' is the count of words that have one other homophone and so on.  It is evident the switch to visemes causes more homophones particularly large numbers of high-multiplicity homophones.  This effect can also be seen in the dictionary size (right of Figure 4).  Homophones cause dictionary entries to merge so the visual dictionary is smaller than the acoustic one.
		\begin{figure}[!htb]
        \centering
		\minipage{0.5\textwidth}
			\centering
			\includegraphics[width=0.85\linewidth]{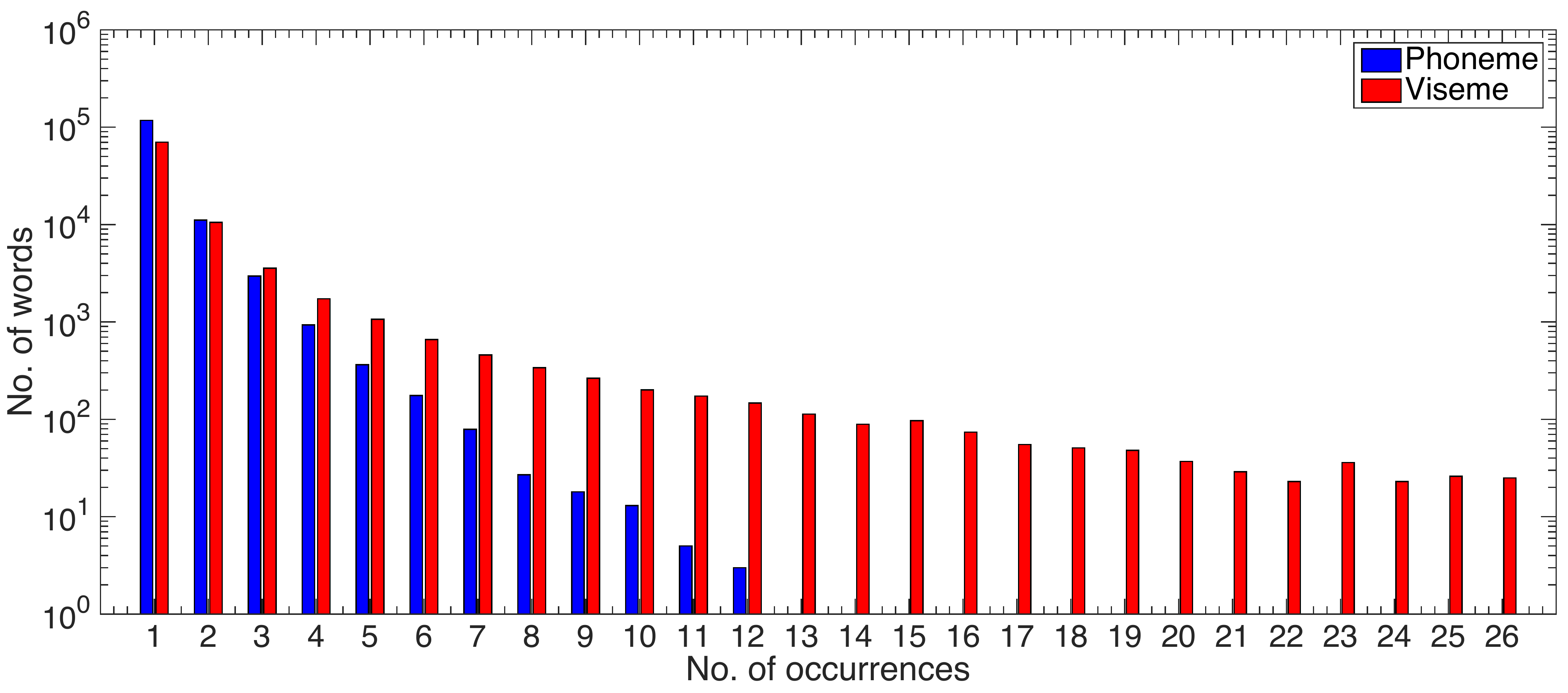} 
			\endminipage\hfill
			\minipage{0.5\textwidth}
			\centering
			\includegraphics[width=\linewidth]{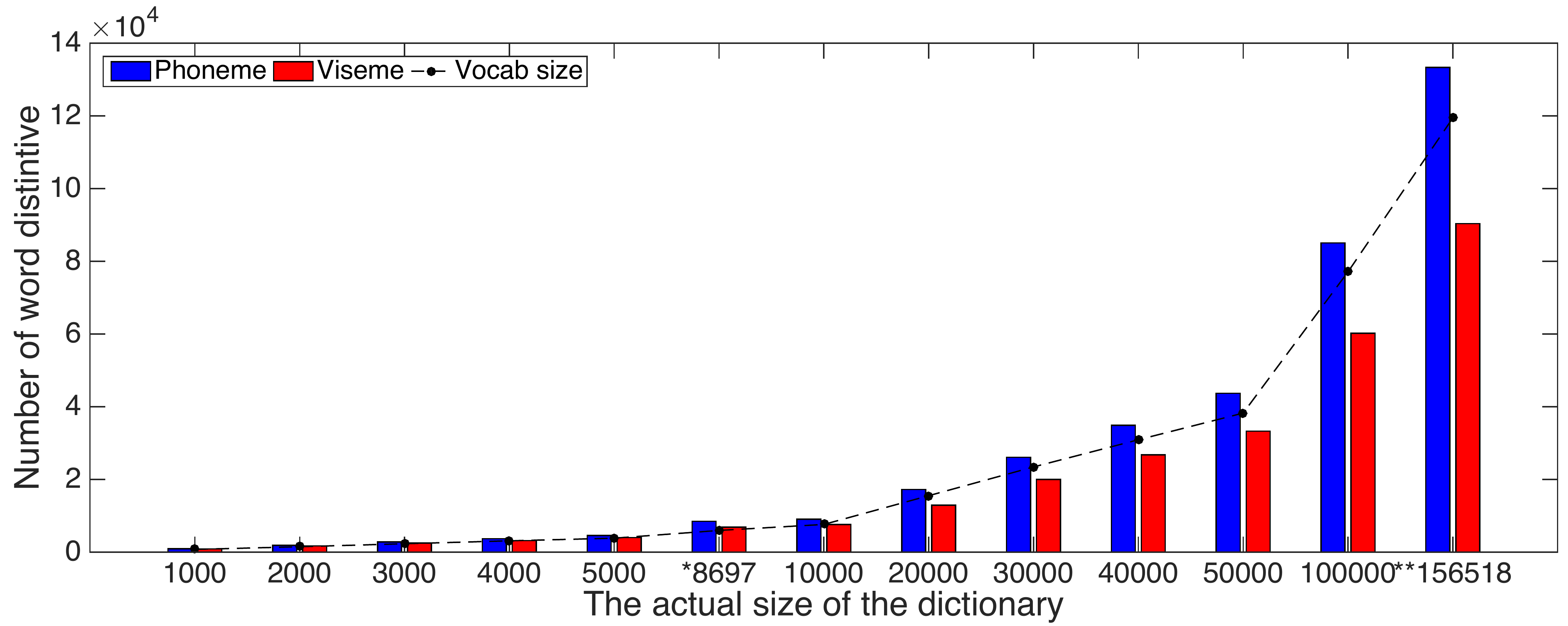} 
			\endminipage
            \label{Fig:Occur}
			\caption{Frequency of duplicated pronunciation in TCD-TIMIT dictionary (left) and vocabulary size (right) for both phoneme and viseme units.}
		\end{figure}

\section{Experiment methodology}
\subsection{Data and Benchmarks}
\label{sec:data}
We use the TCD-TIMIT \cite{TCDTIMIT} corpus containing $59$ volunteer speakers. We chose this dataset because it is the largest vocabulary audio-visual speech corpus available in the public domain. The WFST operates on a vocabulary of almost $6,000$ words from a dictionary of $160,000$ entries.  This dataset provides lists of non-overlapping utterances for training and evaluation in two scenarios: speaker-dependent (SD) and speaker-independent (SI). In the SD scenario, visual models are trained on $3,752$ utterances and evaluated on $1,736$ utterances. Whereas in the SI experiment, $3,822$ utterances from $39$ talkers are in the training set and we evaluate on the remaining $17$ talkers containing a total $1,666$ utterances.  

The TCD-TIMIT release includes a baseline viseme accuracy for both speaker dependent and speaker independent settings using the Neti visemes \cite{neti2000audio} used here. The best viseme accuracy of recognizing $12$ viseme units reported on TCD-TIMIT is $34.77\%$ in speaker independent tests and $34.54\%$ on speaker dependent tests. The context independent viseme models (referred to as mono-viseme in the paper) were trained on $44$-coefficient DCT feature with $4$-state HMMs and $20$ Gaussian mixtures per state.

\section{Results}
\subsection{Viseme-based lipreading experiment}
One fundamental measure of the performance of an automatic lipreading system is viseme accuracy. Since the viseme recognizer requires no dictionary or language model, it is quicker to build and optimise.

Table~\ref{tab:visemeaccuracy} lists the accuracies achieved with our viseme based lipreading system. In comparison to the viseme accuracies benchmarked with the TCD-TIMIT corpus, our best SD viseme accuracy is $46.61\%$ with Eigenlips, compared to $34.54\%$, an improvement of $12.07\%$. Our best SI viseme accuracy is $44.61\%$ which improves on the benchmark $34.77\%$ by $10.16\%$, again with the Eigenlips features. 

\begin{table}[h]
\centering
\caption{Viseme-based lipreading accuracy (\%).}
\label{TB:Beseline-wordacc}
\begin{adjustbox}{width=0.7\columnwidth}
\begin{tabular}{|l|c|c|c|c|c|}
\hline 
\multirow{2}{*}{Model} & \multirow{2}{*}{Feature} & \multicolumn{2}{l|}{Viseme accuracy (\%)} & \multicolumn{2}{l|}{Word accuracy (\%)} \\ \cline{3-6} 
                       &                          & SD                & SI                    & SD               & SI                   \\ \hline
CD-GMM + SAT  &    \multirow{2}{*}{DCT}                  & 44.66  & 42.48   & 14.37  & 10.47  \\ 
CD-DNN &                            & 43.67  & 38.00  & 23.89  & 9.17               \\ \hline 
CD-GMM + SAT & \multirow{2}{*}{Eigenlips}  & 45.59  & 44.61  & 16.71             & 12.15      \\ 
CD-DNN &                            & 46.61  & 44.60  & 33.06 & 19.15           \\ \hline

\end{tabular}
\label{tab:visemeaccuracy}
\end{adjustbox}
\end{table}

Word accuracy achieved with visemes, albeit lower than the viseme accuracy, also shows that Eigenlip features outperform the DCT: we achieved $33.06\%$ in speaker dependent tests, and $19.15\%$ in speaker independent tests. 

\subsection{Phoneme-based lipreading experiment}

Table~\ref{tab:phoemeaccuracy} shows the word and phoneme accuracies achieved with our phoneme-based lipreading system. 
This system achieved the most accurate lipreading with a word accuracy of $48.74\%$. It is interesting that with the phoneme recogniser, word accuracy is greater than phoneme accuracy, because in the viseme recogniser, this is vice versa. 

Again, highest accuracy is achieved with Eigenlip features rather than DCT. 
\begin{table}[!ht]
\centering
\caption{Phoneme-based lipreading accuracy(\%).}
\label{TB:Beseline-wordacc}
\begin{adjustbox}{width=0.7\columnwidth}
\begin{tabular}{|l|c|c|c|c|c|}
\hline 
\multirow{2}{*}{Model} & \multirow{2}{*}{Feature} & \multicolumn{2}{l|}{Phoneme accuracy (\%)} & \multicolumn{2}{l|}{Word accuracy (\%)} \\ \cline{3-6} 
                       &                          & SD                & SI                    & SD               & SI                   \\ \hline
CD-GMM + SAT &  \multirow{2}{*}{DCT}     & 28.22 & 27.37 & 21.88  & 17.72               \\ 
CD-DNN  &                                & 29.18 & 28.08 & 37.40  & 33.87               \\ \hline 
CD-GMM + SAT &\multirow{2}{*}{Eigenlips} & 31.14 & 29.59 & 28.79  & 24.57               \\ 
CD-DNN  &                                & 33.44 & 31.10 & \textbf{48.74}  & \textbf{42.97}               \\ \hline

\end{tabular}
\label{tab:phoemeaccuracy}
\end{adjustbox}
\end{table}
One interesting observation apparent in Tables~\ref{tab:visemeaccuracy} and \ref{tab:phoemeaccuracy} is that the introduction of the DNN makes little difference to the unit accuracy but a bigger difference to a word accuracy for both DCT and eignlips features.

\subsection{Discussion}
\begin{figure}[h]
\centering
\includegraphics[scale=0.7]{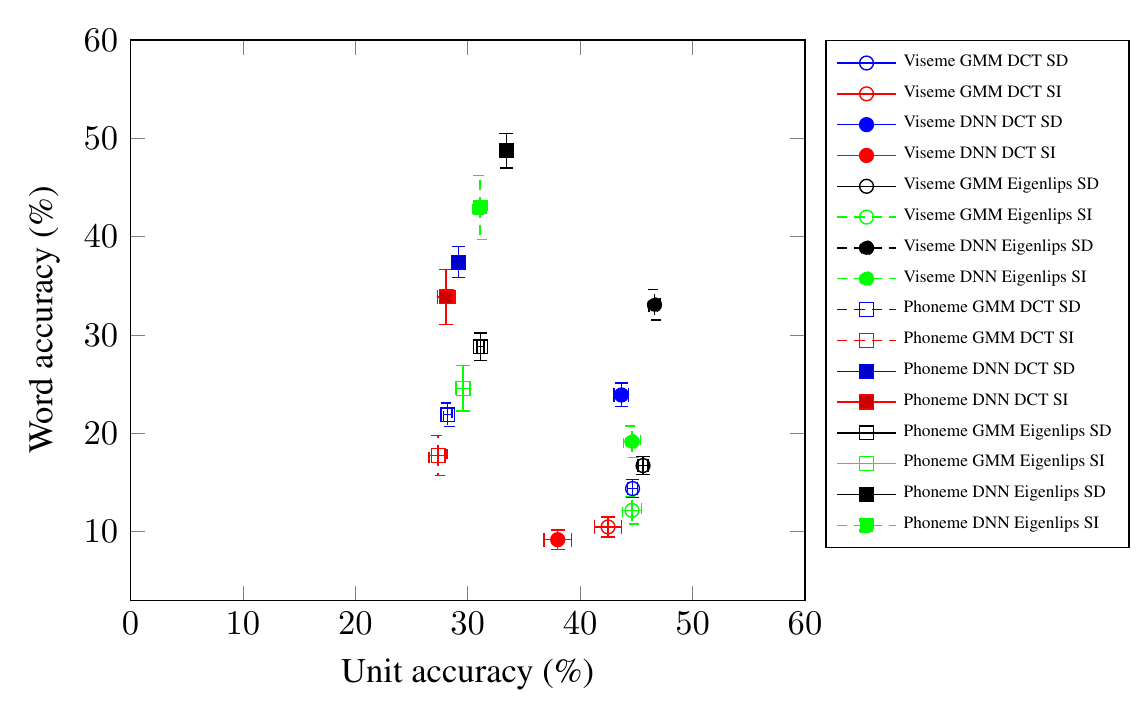}
\caption{Lipreading system performance in GMM system.}
\label{Fig:LipReadingReview}		
\end{figure}

Figure~\ref{Fig:LipReadingReview} plots all of our experimental results comparing unit accuracies (along the $x$-axis) against the word accuracies (on the $y$-axis) along with errorbars showing $\pm1$ standard error.

Figure~\ref{Fig:LipReadingReview} has two clusters: one, in the bottom right, represents the viseme experiments and the other, on the upper left the phonemes.  Here we are representing viseme classifiers with circles (filled represents the DNN, open the GMM) and the phonemes with squares (either filled or open depending on the classifier).  The colours represent the various SI/SD or DCT/Eigenlips combinations.

\begin{figure}[h]
\centering
	\minipage{0.49\textwidth}
	\centering
	\includegraphics[width=0.92\linewidth]{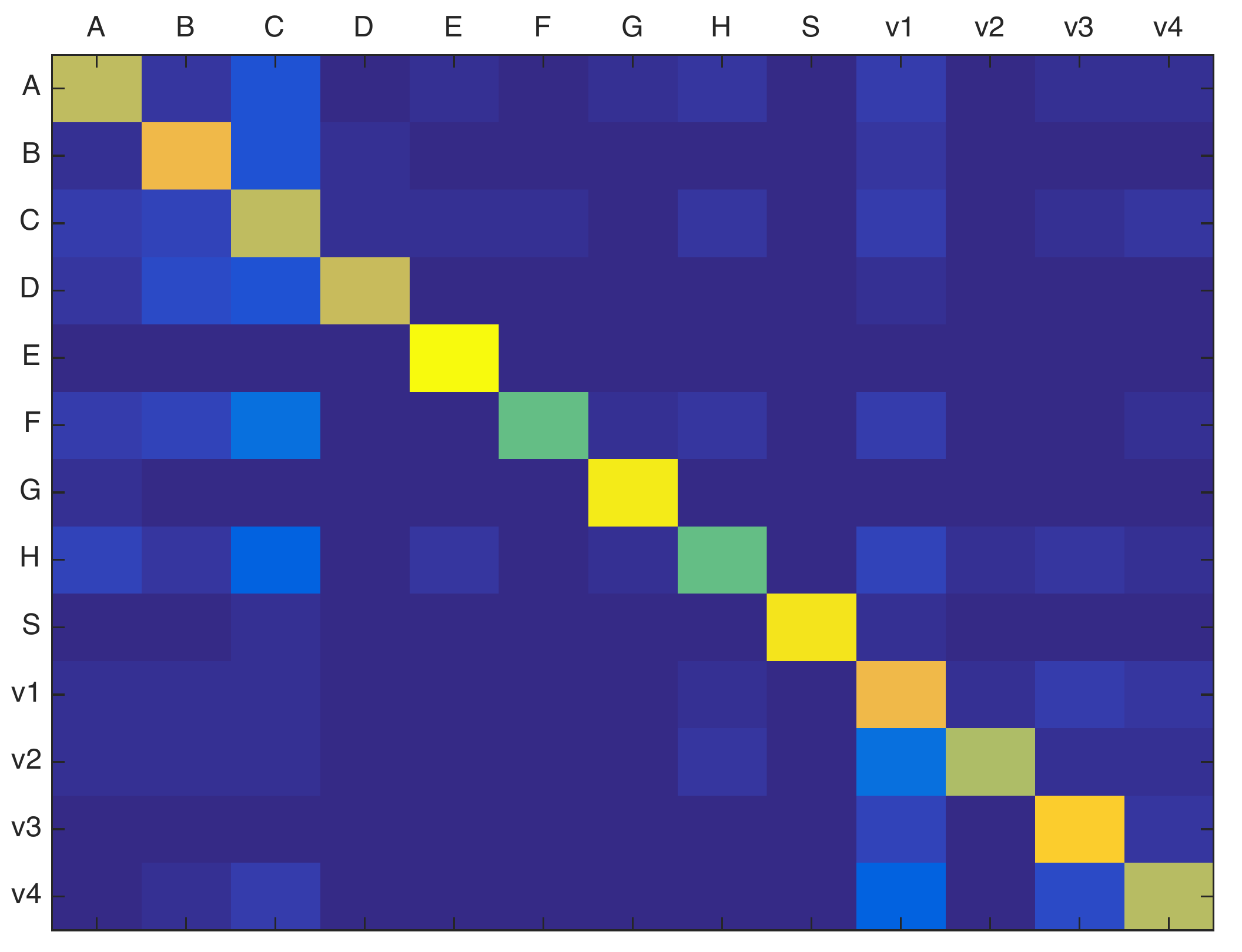} 
    \endminipage\hfill
	\minipage{0.49\textwidth}
	\centering
	\includegraphics[width=0.92\linewidth]{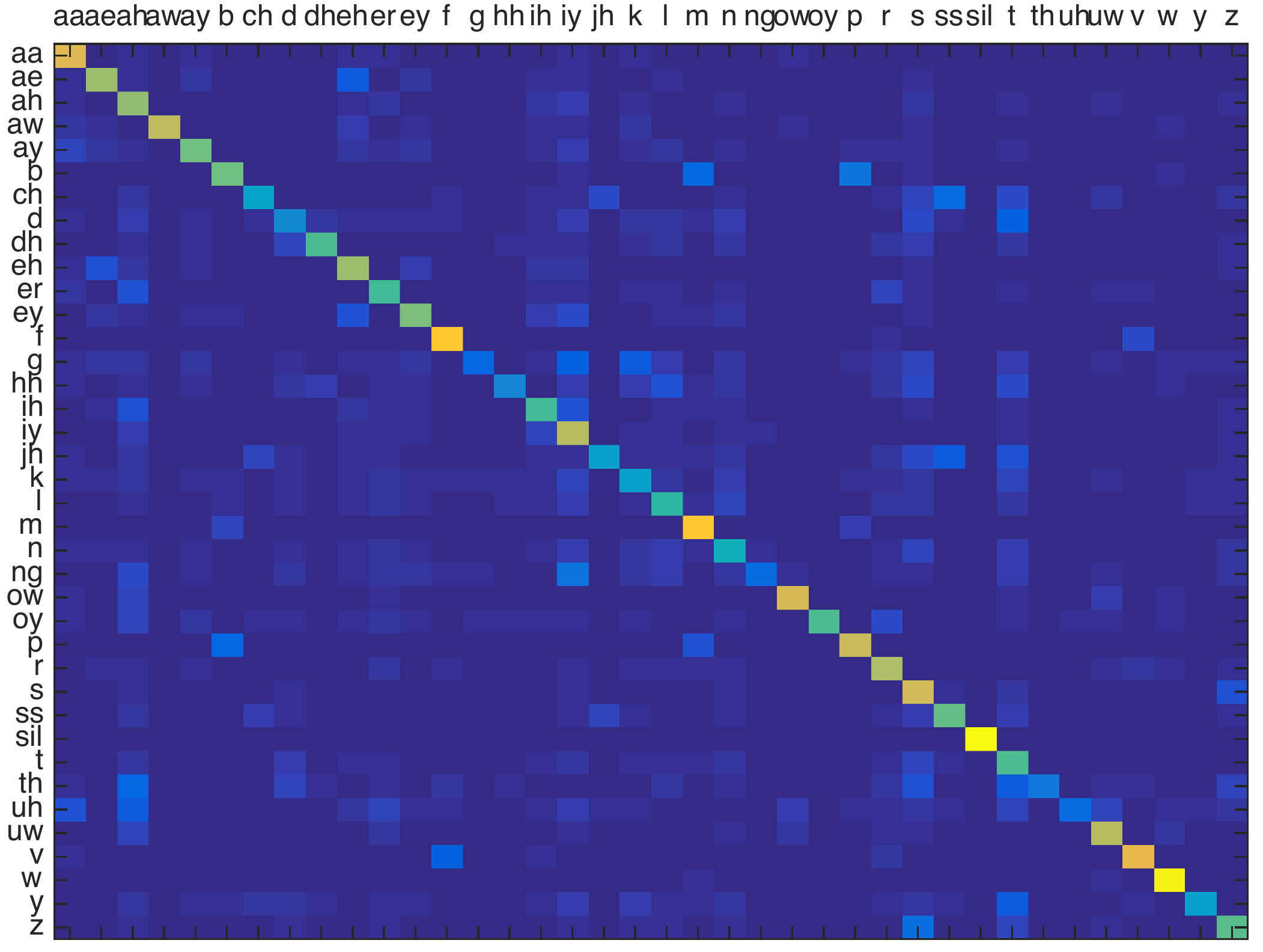} 
	\endminipage
\caption{Comparison of visemes confusion matrix (left) vs phomemes confusion matrix (right).}
\label{fig:confusionmatrixes}
\end{figure}

The phoneme recogniser naturally obtained lower unit accuracy scores because it has three times more phoneme classes than viseme classes ($13$ to $38$ respectively). But this does not mean that phoneme classes have less power to model a visual gesture. This is visualised in the confusion matrices in Figure~\ref{fig:confusionmatrixes} where the colour patterns are consistent between phoneme classes (on the left of Fig~\ref{fig:confusionmatrixes} and between viseme classes on the right of Fig~\ref{fig:confusionmatrixes}). 	  

We note that reducing the set of visual speech units also reduces the discriminant power of the classification model whilst increasing the complexity of pronunciation dictionary by increasing the volume of homophone (homoviseme) words. This suggests that word accuracy of a viseme based system will be less likely to outperform the phoneme based system.

One of the disadvantages of the DNN is that it is not easy to examine to internals of the network to discover from where it is getting its performance.  However there is a clue in the previous observation which is that the DNN appears to make the most difference to word accuracy rather than unit accuracy.  Visual speech is notorious for extensive co-articulation so the implication is that either there are significant differences in the window length between the GMM and the DNN or the DNN is better able to model co-articulation than the GMM.  Although there are variations in the window length, here the GMM has a slightly longer span of $150$ms compared to $110$ms for the DNN, it is latter explanation is the most likely.  In other work \cite{LipreadingIS2017} we were able to use identical features and we also found the DNN superior furthermore we know the DNN to be better able to learn data structured on non-linear manifolds so we believe this is the most likely explanation for the success of the DNN.

One caveat is that we have not optimised the scaling factor used in our language models so there is probably more performance to come when measured as word accuracy.
\section{Conclusion}
One observation we found is that DNN-HMM viseme recognizers can easily overfit to the training observations, this is shown in the performance disparity between SD and SI configurations. It could potentially be interesting to use visemes as an initialisation for phoneme recognition in a hierarchical training method similar to that in \cite{bear2016decoding} in the future. 

We have added more evidence to the argument that phoneme classifiers can outperform those of visemes. Whilst there is still debate about visemes, we can not forget them, but given the evidence showing a significant improvement in word accuracy from the reduction in homophonic words in a pronunciation dictionary, we suggest that phonemes are the current optimal class labels for lipreading.

We have also illustrated the noticeable performance gain by changing visual representation from DCT to Eigenlips. The best word accuracy in this work is $48.74\%$ on SD and $42.97\%$ on SI achieved with the DNN-HMM phoneme unit recognizer trained on Eigenlip features. However, the disadvantage of Eigenlip feature is a learned linear mapping that needs to be trained.

Conventional systems have shown speaker independence to be a challenge, here with a novel DNN-HMM architecture, we have reduced the effect between these arrangements. We speculate that the success of the DNN is likely to do its ability to better model the effects of co-articulation which is a well known bugbear of human and machine lip-readers.

\bibliography{mybib}

\begin{thebibliography}{27}
\providecommand{\natexlab}[1]{#1}
\providecommand{\url}[1]{\texttt{#1}}
\expandafter\ifx\csname urlstyle\endcsname\relax
  \providecommand{\doi}[1]{doi: #1}\else
  \providecommand{\doi}{doi: \begingroup \urlstyle{rm}\Url}\fi

\bibitem[Ahmed et~al.(1974)Ahmed, Natarajan, and Rao]{ahmed1974discrete}
Nasir Ahmed, T\_ Natarajan, and Kamisetty~R Rao.
\newblock Discrete cosine transform.
\newblock \emph{IEEE transactions on Computers}, 100\penalty0 (1):\penalty0
  90--93, 1974.

\bibitem[Almajai et~al.(2016)Almajai, Cox, Harvey, and Lan]{LipreadingSAT2016}
I.~Almajai, S.~Cox, R.~Harvey, and Y.~Lan.
\newblock Improved speaker independent lip reading using speaker adaptive
  training and deep neural networks.
\newblock In \emph{2016 IEEE International Conference on Acoustics, Speech and
  Signal Processing (ICASSP)}, pages 2722--2726, March 2016.
\newblock \doi{10.1109/ICASSP.2016.7472172}.

\bibitem[Association(1999)]{international1999handbook}
International~Phonetic Association.
\newblock \emph{Handbook of the International Phonetic Association: A guide to
  the use of the International Phonetic Alphabet}.
\newblock Cambridge University Press, 1999.

\bibitem[Bear and Harvey(2016)]{bear2016decoding}
Helen~L Bear and Richard Harvey.
\newblock Decoding visemes: Improving machine lip-reading.
\newblock In \emph{Acoustics, Speech and Signal Processing (ICASSP), 2016 IEEE
  International Conference on}, pages 2009--2013. IEEE, 2016.

\bibitem[Bear and Taylor(2017)]{bearTaylor2017}
Helen~L Bear and Sarah Taylor.
\newblock Visual speech recognition: aligning terminologies for better
  understanding.
\newblock In \emph{British Machine Vision Conference (BMVC), Deep Learning for
  Machine Lip Reading workshop}, 2017.

\bibitem[Bear et~al.(2014)Bear, Harvey, Theobald, and Lan]{bear2014phoneme}
Helen~L Bear, Richard~W Harvey, Barry-John Theobald, and Yuxuan Lan.
\newblock Which phoneme-to-viseme maps best improve visual-only computer
  lip-reading?
\newblock In \emph{Advances in Visual Computing}, pages 230--239. Springer,
  2014.
\newblock \doi{10.1007/978-3-319-14364-4_22}.

\bibitem[Bregler and Konig(1994)]{bregler1994eigenlips}
Christoph Bregler and Yochai Konig.
\newblock " eigenlips" for robust speech recognition.
\newblock In \emph{Acoustics, Speech, and Signal Processing, 1994. ICASSP-94.,
  1994 IEEE International Conference on}, volume~2, pages II--669. IEEE, 1994.

\bibitem[Cox et~al.(2008)Cox, Harvey, Lan, Newman, and
  Theobald]{cox2008challenge}
Stephen~J Cox, Richard~W Harvey, Yuxuan Lan, Jacob~L Newman, and Barry-John
  Theobald.
\newblock The challenge of multispeaker lip-reading.
\newblock In \emph{AVSP}, pages 179--184, 2008.

\bibitem[Fisher(1936)]{fisher1936use}
Ronald~A Fisher.
\newblock The use of multiple measurements in taxonomic problems.
\newblock \emph{Annals of human genetics}, 7\penalty0 (2):\penalty0 179--188,
  1936.

\bibitem[Gales(1998)]{Gales1998}
M.J.F. Gales.
\newblock Maximum likelihood linear transformations for {HMM-based} speech
  recognition.
\newblock \emph{Computer Speech \& Language}, 12\penalty0 (2):\penalty0 75 --
  98, 1998.
\newblock ISSN 0885-2308.

\bibitem[Harte and Gillen(2015)]{TCDTIMIT}
N.~Harte and E.~Gillen.
\newblock {TCD-TIMIT}: An audio-visual corpus of continuous speech.
\newblock \emph{IEEE Transactions on Multimedia}, 17\penalty0 (5):\penalty0
  603--615, May 2015.
\newblock ISSN 1520-9210.
\newblock \doi{10.1109/TMM.2015.2407694}.

\bibitem[Hecht-Nielsen et~al.(1988)]{hecht1988theory}
Robert Hecht-Nielsen et~al.
\newblock Theory of the backpropagation neural network.
\newblock \emph{Neural Networks}, 1\penalty0 (Supplement-1):\penalty0 445--448,
  1988.

\bibitem[Hinton et~al.(2012)Hinton, Deng, Yu, Dahl, Mohamed, Jaitly, Senior,
  Vanhoucke, Nguyen, Sainath, et~al.]{Hinton2012}
Geoffrey Hinton, Li~Deng, Dong Yu, George~E Dahl, Abdel-rahman Mohamed, Navdeep
  Jaitly, Andrew Senior, Vincent Vanhoucke, Patrick Nguyen, Tara~N Sainath,
  et~al.
\newblock Deep neural networks for acoustic modeling in speech recognition: The
  shared views of four research groups.
\newblock \emph{Signal Processing Magazine, IEEE}, 29\penalty0 (6):\penalty0
  82--97, 2012.

\bibitem[Howell et~al.(2016)Howell, Cox, and Theobald]{howell2016visual}
Dominic Howell, Stephen Cox, and Barry Theobald.
\newblock Visual units and confusion modelling for automatic lip-reading.
\newblock \emph{Image and Vision Computing}, 51:\penalty0 1--12, 2016.

\bibitem[Kirby et~al.(1993)Kirby, Weisser, and Dangelmayr]{KIRBY199363}
M~Kirby, F~Weisser, and G~Dangelmayr.
\newblock A model problem in the representation of digital image sequences.
\newblock \emph{Pattern Recognition}, 26\penalty0 (1):\penalty0 63 -- 73, 1993.
\newblock ISSN 0031-3203.
\newblock \doi{http://dx.doi.org/10.1016/0031-3203(93)90088-E}.
\newblock URL
  \url{http://www.sciencedirect.com/science/article/pii/003132039390088E}.

\bibitem[Mao and Jain(1995)]{mao1995artificial}
Jianchang Mao and Anil~K Jain.
\newblock Artificial neural networks for feature extraction and multivariate
  data projection.
\newblock \emph{IEEE transactions on neural networks}, 6\penalty0 (2):\penalty0
  296--317, 1995.

\bibitem[Matthews et~al.(2002)Matthews, Cootes, Bangham, Cox, and
  Harvey]{Matthews02}
I~Matthews, TF~Cootes, JA~Bangham, SJ~Cox, and RW~Harvey.
\newblock Extraction of visual features for lipreading.
\newblock \emph{IEEE transactions on Pattern Analysis and Machine
  Intelligence}, 24\penalty0 (2):\penalty0 198--213, 2002.
\newblock ISSN 0162-8828.
\newblock \doi{10.1109/34.982900}.

\bibitem[Neti et~al.(2000{\natexlab{a}})Neti, Potamianos, Luettin, Matthews,
  Glotin, Vergyri, Sison, Mashari, and Zhou]{neti}
Neti, Potamianos, Luettin, Matthews, Glotin, Vergyri, Sison, Mashari, and Zhou.
\newblock Audio-visual speech recognition, 2000{\natexlab{a}}.
\newblock Technical report, Center for Language and Speech Processing, The
  Johns Hopkins University, Baltimore.

\bibitem[Neti et~al.(2000{\natexlab{b}})Neti, Potamianos, Luettin, Matthews,
  Glotin, Vergyri, Sison, and Mashari]{neti2000audio}
Chalapathy Neti, Gerasimos Potamianos, Juergen Luettin, Iain Matthews, Herve
  Glotin, Dimitra Vergyri, June Sison, and Azad Mashari.
\newblock Audio visual speech recognition.
\newblock Technical report, IDIAP, 2000{\natexlab{b}}.

\bibitem[Potamianos et~al.(2001)Potamianos, Luettin, and Neti]{Potamianos2001}
Gerasimos Potamianos, Juergen Luettin, and Chalapathy Neti.
\newblock Hierarchical discriminant features for audio-visual {LVCSR}.
\newblock In \emph{Acoustics, Speech, and Signal Processing, 2001.
  Proceedings.(ICASSP'01). 2001 IEEE International Conference on}, volume~1,
  pages 165--168. IEEE, 2001.

\bibitem[Povey and Saon(2006)]{fMLLR2006}
Daniel Povey and George Saon.
\newblock Feature and model space speaker adaptation with full covariance
  {Gaussians}.
\newblock In \emph{{INTERSPEECH} 2006 - ICSLP, Ninth International Conference
  on Spoken Language Processing, Pittsburgh, PA, USA, September 17-21, 2006},
  2006.

\bibitem[Povey et~al.(2011)Povey, Ghoshal, Boulianne, Goel, Hannemann, Qian,
  Schwarz, and Stemmer]{Povey11}
Daniel Povey, Arnab Ghoshal, Gilles Boulianne, Nagendra Goel, Mirko Hannemann,
  Yanmin Qian, Petr Schwarz, and Georg Stemmer.
\newblock The {Kaldi} speech recognition toolkit.
\newblock In \emph{In IEEE 2011 workshop}, 2011.

\bibitem[Rabiner and Juang(1986)]{rabiner1986introduction}
Lawrence Rabiner and B~Juang.
\newblock An introduction to hidden markov models.
\newblock \emph{ieee assp magazine}, 3\penalty0 (1):\penalty0 4--16, 1986.

\bibitem[Thangthai and Harvey(2017)]{LipreadingIS2017}
Kwanchiva Thangthai and Richard Harvey.
\newblock Improving computer lipreading via dnn sequence discriminative
  training techniques.
\newblock In \emph{INTERSPEECH 2017 (to be published)}, 2017.

\bibitem[Viterbi(1967)]{viterbi1967error}
Andrew Viterbi.
\newblock Error bounds for convolutional codes and an asymptotically optimum
  decoding algorithm.
\newblock \emph{IEEE transactions on Information Theory}, 13\penalty0
  (2):\penalty0 260--269, 1967.

\bibitem[Wold et~al.(1987)Wold, Esbensen, and Geladi]{wold1987principal}
Svante Wold, Kim Esbensen, and Paul Geladi.
\newblock Principal component analysis.
\newblock \emph{Chemometrics and intelligent laboratory systems}, 2\penalty0
  (1-3):\penalty0 37--52, 1987.

\bibitem[Xilinx(2002)]{2ddct2}
Inc. Xilinx.
\newblock 2d discrete cosine transform (dct) v2.0 logicore product
  specification, 2002.

\end{thebibliography}
\end{document}